\documentclass{article} 
\usepackage{nips12submit_e,times}
\usepackage{natbib}
\usepackage{amsmath}
\usepackage{amsfonts}
\usepackage{graphicx}

\title{Joint Training of Deep Boltzmann Machines for Classification}

\author{Ian J. Goodfellow \And Aaron Courville\\Universit\'e de Montr\'eal \And Yoshua Bengio}

\nipsfinalcopy 

\begin{document}

\maketitle

\begin{abstract}
We introduce a new method for training deep Boltzmann machines jointly.
Prior methods of training DBMs require an initial learning pass that trains
the model greedily, one layer at a time, or do not
perform well on classification tasks. In our approach, we train all
layers of the DBM simultaneously, using a novel training procedure
called {\em multi-prediction training}. The resulting model can either
be interpreted as a single generative model trained to maximize a variational
approximation to the generalized pseudolikelihood, or as a family of recurrent
networks that share parameters and may be approximately averaged together
using a novel technique we call the {\em multi-inference trick}. We show
that our approach performs competitively for classification and outperforms
previous methods in terms of accuracy of approximate inference and classification
with missing inputs.
\end{abstract}

\section{Deep Boltzmann machines}

A deep Boltzmann machine \citep{Salakhutdinov2009} is a probabilistic model
consisting of many layers of random variables, most of
which are latent. Typically, a DBM contains a set of $D$ input features $v$ that are called the {\em visible units}
because they are always observed during both training and evaluation.
The DBM is usually applied to classification problems and thus often represents the class label with a one-of-$k$ code
in the form of a discrete-valued label unit $y$. $y$  is observed (on examples for which it is
available) during training.
The DBM also contains several latent variables that are never observed. These {\em hidden units} are usually organized into $L$ layers
$h^{(i)}$ of size $N_i, i=1,\dots,L$, with each unit in a layer conditionally independent of the other units in the layer given
the neighboring layers. These conditional independence properties allow fast Gibbs sampling because an entire layer of
units can be sampled at a time. Likewise, mean field inference with fixed point equations is fast because each fixed
point equation gives a solution to roughly half of the variational parameters. Inference proceeds by alternating
between updating all of the even numbered layers and updating all of the odd numbered layers.

A DBM defines a probability distribution by exponentiating and normalizing an energy function
\[
P(v,h,y) = \frac{1}{Z} \exp\left( -E(v,h,y) \right) \]
where
\[
Z = \sum_{v',h',y'} \exp \left( -E(v', h', y') \right). \]

$Z$, the partition function, is intractable, due to the summation over all possible states. Maximum likelihood
learning requires computing the gradient of $\log Z$. Fortunately, the gradient can be estimated using an MCMC
procedure \citep{Younes1999,Tieleman08}. Block Gibbs sampling of the layers makes this procedure efficient.

The structure of the interactions in $h$ determines whether further approximations are necessary. In the pathological
case where every element of $h$ is conditionally independent of the others given the visible units, the DBM is simply
an RBM and $\log Z$ is the only intractable term of the log likelihood.
In the general case, interactions between different elements of $h$ render the posterior $P(h \mid v, y)$ intractable.
\citet{Salakhutdinov2009} overcome this
by maximizing the lower bound on the log likelihood given by the mean field approximation to
the posterior rather than maximizing the log likelihood itself. Again, block mean field inference over the layers makes
this procedure efficient.

An interesting property of the DBM is that the training procedure thus involves {\em feedback connections} between the
layers. Consider the simple DBM consisting entirely of binary-valued units, with the energy function
\[ E(v,h) = - v^T W^{(1)} h^{(1)}  -h^{(1)T} W^{(2)} h^{(2)}. \]
Approximate inference in this model involves repeatedly applying two fixed-point update equations to solve for the
mean field approximation to the posterior. Essentially it involves running a recurrent net in order to obtain approximate
expectations of the latent variables.

Beyond their theoretical appeal as a deep model that admits simultaneous training of all components using a generative
cost, DBMs have achieved excellent performance in practice. When they were first introduced, DBMs set the state of the
art on the permutation-invariant\footnote{By permutation-invariant, we
mean that permuting all of the input pixels prior to learning the network should not cause a change in performance, so
using synthetic image distortions or convolution to engineer knowledge about the structure of the images into the system
is not allowed.} version of the MNIST handwritten digit recognition task at 0.95\%.

Recently, new techniques were used in conjunction with DBM pretraining to set a new state
of the art of 0.79\% test error \citep{Hinton-et-al-arxiv2012}.

\section{The joint training problem}

\begin{figure}
\center
\includegraphics[width=\textwidth]{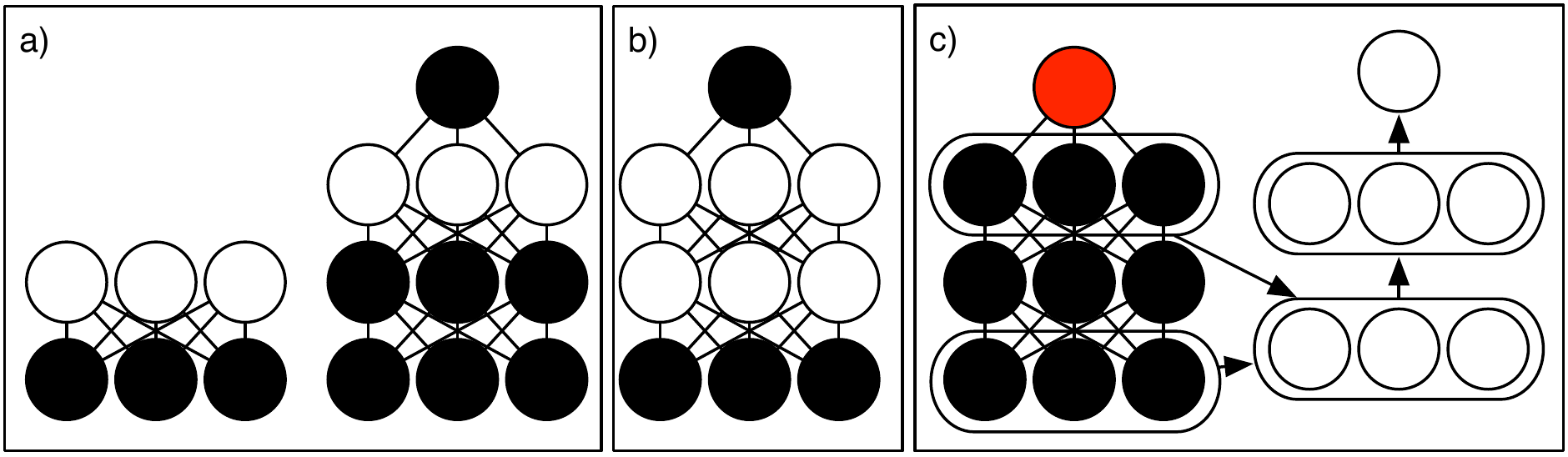}
\caption{The training procedure employed by \citet{Salakhutdinov2009} on MNIST.
a) An RBM comprising $v$ and $h^{(1)}$ is trained to maximize the log likelihood of $v$ using CD.
Next, another RBM is trained with CD,
using $y$ and samples of $h^{(1)}$ conditioned on $v$ as observed data, and $h^{(2)}$ as hidden units.
b) The two RBMs are stitched together to form one DBM over $v$, $h^{(1)}$, $h^{(2)}$, and $y$.
This DBM is trained to maximize the log likelihood of $v$ and $y$ using PCD.
c) $y$ is deleted from the model. An extra MLP is built on top of $v$ and the mean field expectations
of $h^(1)$ and $h^(2)$. The parameters of the DBM are frozen, and the parameters of the MLP are initialized
based on the DBM parameters, then trained via nonlinear conjugate gradient descent to predict $y$ from $v$
and the mean field features.
}
\label{standard_dbm_training}
\end{figure}

Unfortunately, it is not possible to train a deep Boltzmann machine using only the variational
bound and approximate gradient described above. See \citep{GoodfellowTPAMI_ToAppear} for an example
of a DBM that has failed to learn using the naive training algorithm. \citet{Salakhutdinov2009} found
that for CD / PCD to work, the DBM must be initialized by training one layer at a time.
After each layer is trained as an RBM, the RBMs can be
modified slightly, assembled into a DBM, and the DBM may be trained with PCD learning rule described above.
In order to achieve good classification results, an MLP designed specifically to predict
$y$ from $v$ must be trained on top of the DBM model. Simply running mean field inference to predict $y$
given $v$ in the DBM model does not work nearly as well. See figure \ref{standard_dbm_training} for a
graphical description of the training procedure used by \cite{Salakhutdinov2009}.

In this paper, we propose a method that enables the deep Boltzmann machine to be jointly trained, and
to achieve excellent performance as a classifier without an additional classification-specific extension
of the model. The standard approach requires training $L+2$ different models using $L+2$ different objective
functions, and does not yield a single model that excels at answering all queries. Our approach requires training
only one model with only one objective function, and the resulting model outperforms previous approaches at answering
all kinds of queries (classification, classification with missing inputs, predicting arbitrary subsets of variables
given arbitrary subsets of variables).

\section{Motivation}

There are numerous reasons to prefer a single-model, single-training stage approach to deep Boltzmann machine
learning:

\begin{enumerate}
\item {\bf Optimization} 
As a greedy optimization procedure, layerwise training may be suboptimal.
Small-scale experimental work has demonstrated this to be the case for
deep belief networks \citep{Arnold+Ollivier-arxiv2012}.

In general, for layerwise training to be optimal, the training procedure for
each layer must take into account the influence that the deeper layers will
provide. The standard training layerwise procedure simply does not attempt to be optimal.

The procedures used by \citet{LeRoux-Bengio-2008, Arnold+Ollivier-arxiv2012}
make an optimistic
assumption that the deeper layers will be able to implement the best possible
prior on the current layer's hidden units.
This approach is not immediately applicable to Boltzmann machines because
it is specified in terms of learning the parameters of
$P(h^{(i-1)} | h^{(i)})$ assuming that the parameters of the $P(h^{(i)})$ will be
set optimally later. In a DBM the symmetrical nature of the interactions between
units means that these two distributions share parameters, so it is not possible to
set the parameters of the one distribution, leave them fixed for the remainder of
learning, and then set the parameters of the other distribution.
Moreover, model architectures incorporating design features
such as sparse connections, pooling, or factored multilinear interactions make
it difficult to predict how best to structure one layer's hidden units in order
for the next layer to make good use of them.

\item {\bf Probabilistic modeling}
Using multiple models and having some models specialized for exactly one task (like
predicting $y$ from $v$) loses some of the benefit of probabilistic modeling. If
we have one model that excels at all tasks, we can use inference in this model to
answer arbitrary queries, perform classification with missing inputs, and so on.

\item {\bf Simplicity }
Needing to implement multiple models and training stages makes the cost of developing
software with DBMs greater, and makes using them more cumbersome. Beyond the practical
considerations, it can be difficult to monitor training and tell what kind of results
during layerwise DBM pretraining will correspond to good classification accuracy later.
Our joint training procedure allows the user to monitor the model's ability of interest
(usually ability to classify $y$ given $v$) from the very start of training.
\end{enumerate}

\section{Multi-Prediction Training}

\begin{figure}
\center
\includegraphics[width=0.5 \textwidth]{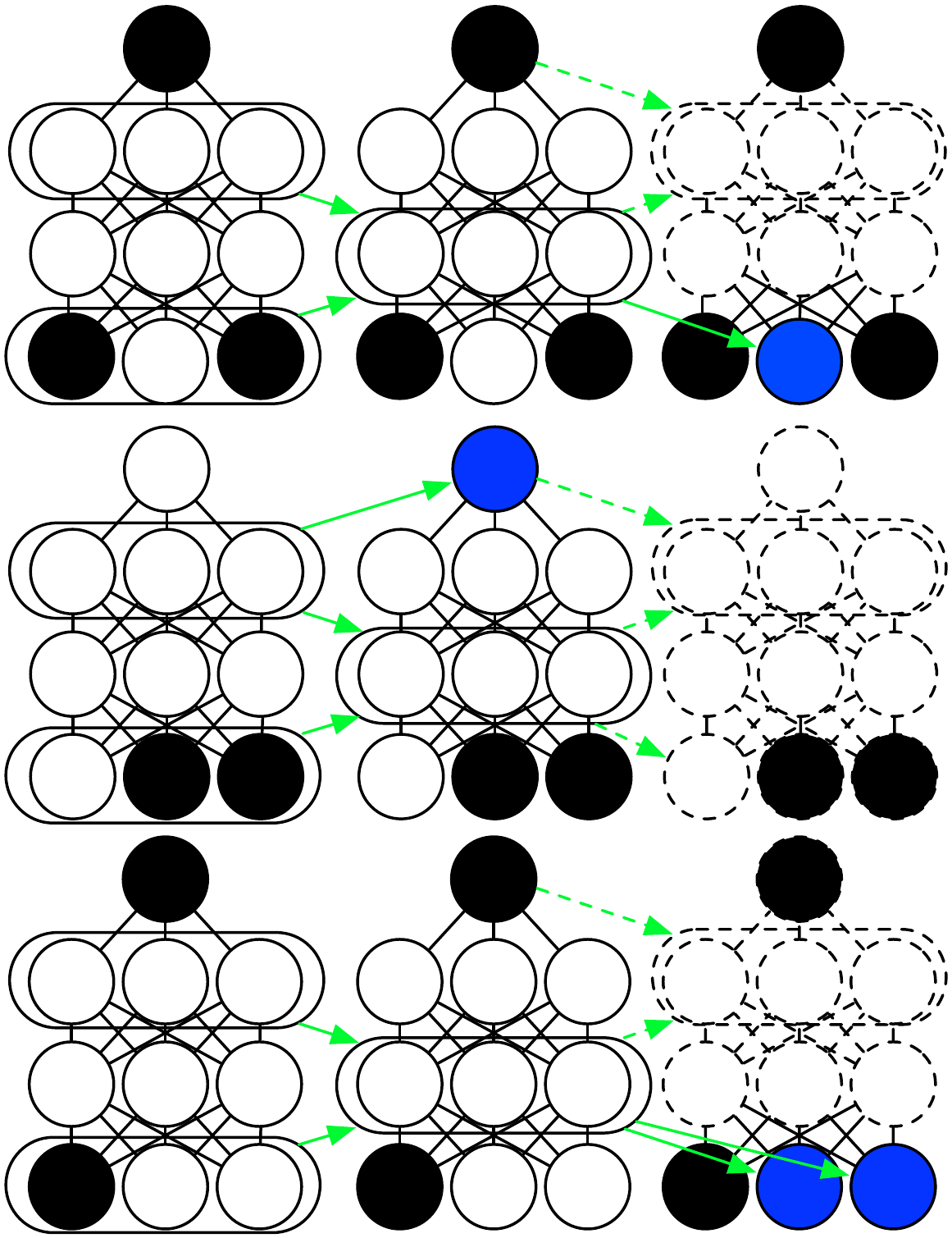}
\caption{
{\em Graphical description of multi-prediction training}. During MP training, we sample
examples $(v, y)$ from the training set. For each example, we choose one subset of variables
(uniformly at random) to serve as the input to an inference problem, and use the complement
of this subset as targets for training. We then run mean field inference and backpropagate
the prediction error through the entire computational graph of the inference process. Here,
we depict the process for three different examples, one example per row. We use black circles
to indicate observed data and blue circles to indicate prediction targets. The green arrows
show the flow of computation through the mean field graph. Each column indicates another time
step in the mean field inference algorithm. Dotted lines indicate quantities that are not used
for this instantiation of the problem, but would be used if we ran another iteration of mean
field. Here, we show only one iteration of mean field.
To work well, MP training should be run with 5-10 iterations of mean field.
}
\label{mpt}
\end{figure}

\begin{figure}
\center
\includegraphics[width=0.5\textwidth]{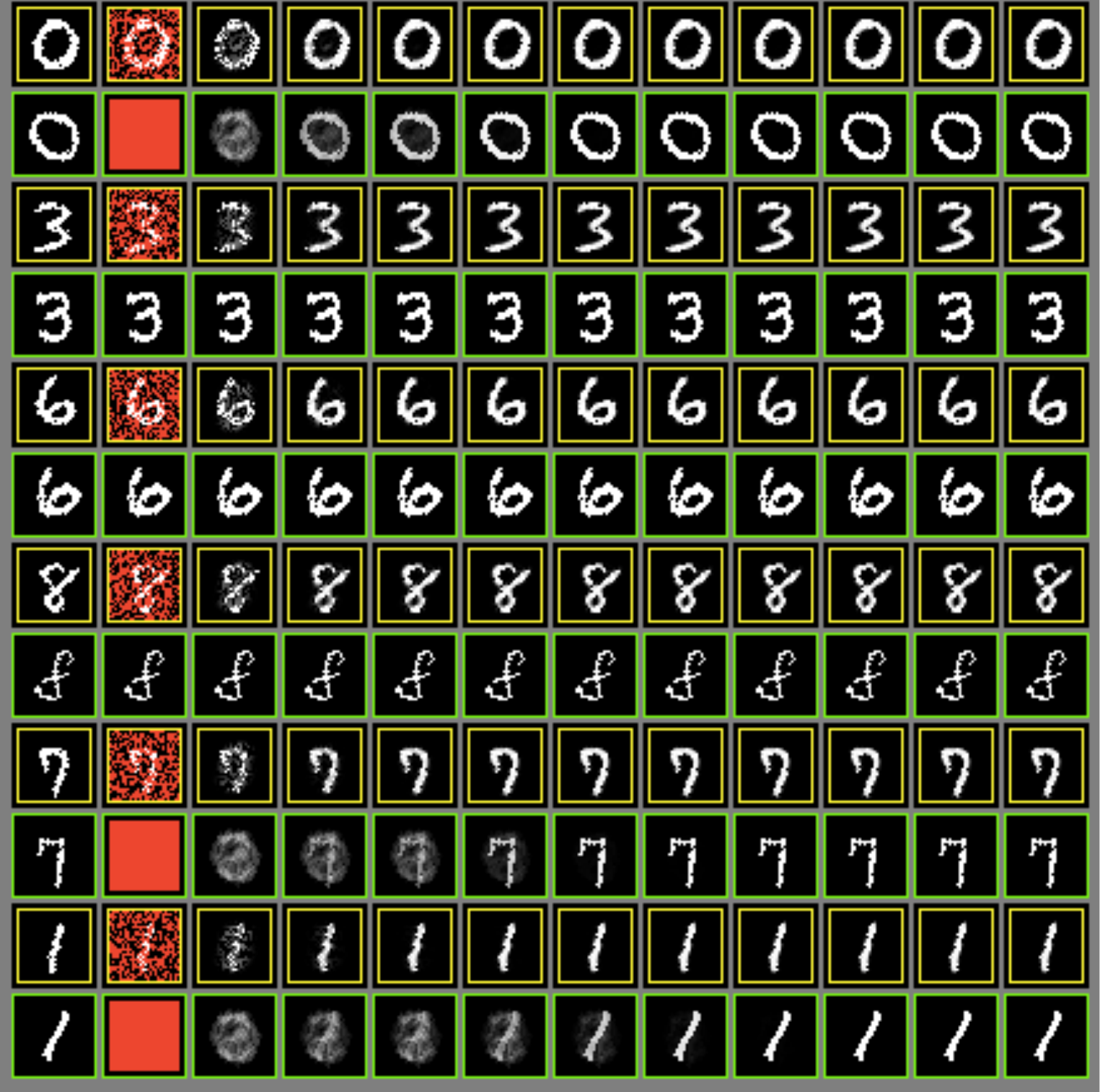}
\caption{Mean field inference applied to MNIST digits.
The first column shows the true digits.
The second column shows pixels of the digits to be masked out, with red pixels indicating the region to
be witheld from the input to the DBM.
Yellow-boxed rows show input pixels. Green-boxed rows represent the class variables.
The subsequent columns show the DBM incrementally predicting the missing variables, with each column
being one iteration of mean field. On rows where the green-boxed class variable was masked out, the
uncertainty over the class is represented by displaying a weighted average of templates for the 10 different
classes.}
\label{animate_inpainting}
\end{figure}

Our proposed approach is to directly train the DBM to be good at solving all possible
variational inference problems. We call this {\em multi-prediction training} because
the procedure involves training the model to predict any subset of variables given the
complement of that subset of variables.

Specifically, we use stochastic gradient descent on the {\em multi-prediction} (MP) objective function

\[ J(v, \theta) = - \sum_i \log Q^*_{v_{-S_i}} ( v_{S_i} ) \]

where $S$ is a sequence of subsets of the possible indices of $v$ and
\[ Q^*_{v_{-S_i}}(v_{S_i}, h) = \text{argmin}_Q D_{KL} \left( Q(v_{S_i}, h ) \Vert P( v_{S_i}, h \mid v_{-S_i} ) \right) . \]

In other words, the criterion for a single example $v$ is a sum of several terms, with term $i$ measuring the model's
ability to predict a subset of the inputs, $v_{S_i}$, given the remainder of the inputs, $v_{-S_i}$.

During SGD training, we sample minibatches of values of $v$ and $S_i$. Sampling an $S_i$ uniformly simply requires sampling one bit (1 with probability 0.5) for each variable, to determine whether that variable should be an input to the
inference procedure or a prediction target.

In this paper, $Q$ is constrained to be factorial, though one could design
model families for which it makes sense to use richer structure in $Q$.
In order to accomplish the minimization, we instantiate a recurrent net that repreatedly runs the mean field
fixed point equations, and backpropagrate the gradient of $J$ through the entire recurrent net.

See Fig. \ref{mpt} for a graphical description of this training procedure, and Fig. \ref{animate_inpainting}
for an example of the inference procedure run on MNIST digits.

\section{The Multi-Inference Trick}

\begin{figure}
\center
\includegraphics[width=0.5 \textwidth]{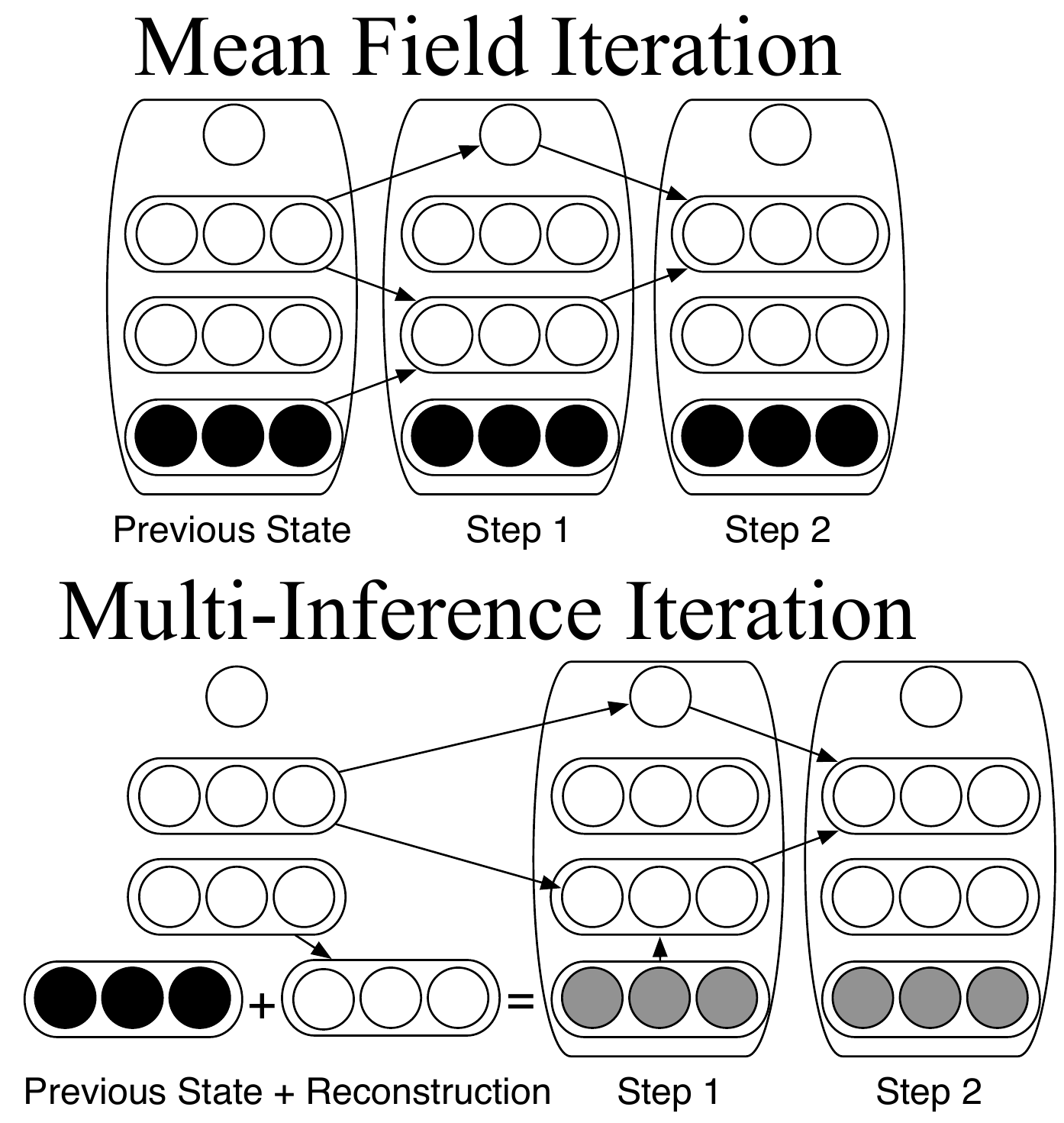}
\caption{
{\em Graphical description of the multi-inference trick}. Consider the problem of estimating $y$
given $v$.
A mean field iteration consists of first applying a mean field update to $h^{(1)}$ and $y$, then
applying one to $h^{(2)}$.
When using the multi-inference trick, we start the iteration by first computing $r$ as the mean
field update $v$ would receive if it were not observed. We then use $0.5 (r+v)$ in place of $v$
and run a regular mean field iteration.
}
\label{mi}
\end{figure}

\begin{figure}
\center
\includegraphics[width=0.5 \textwidth]{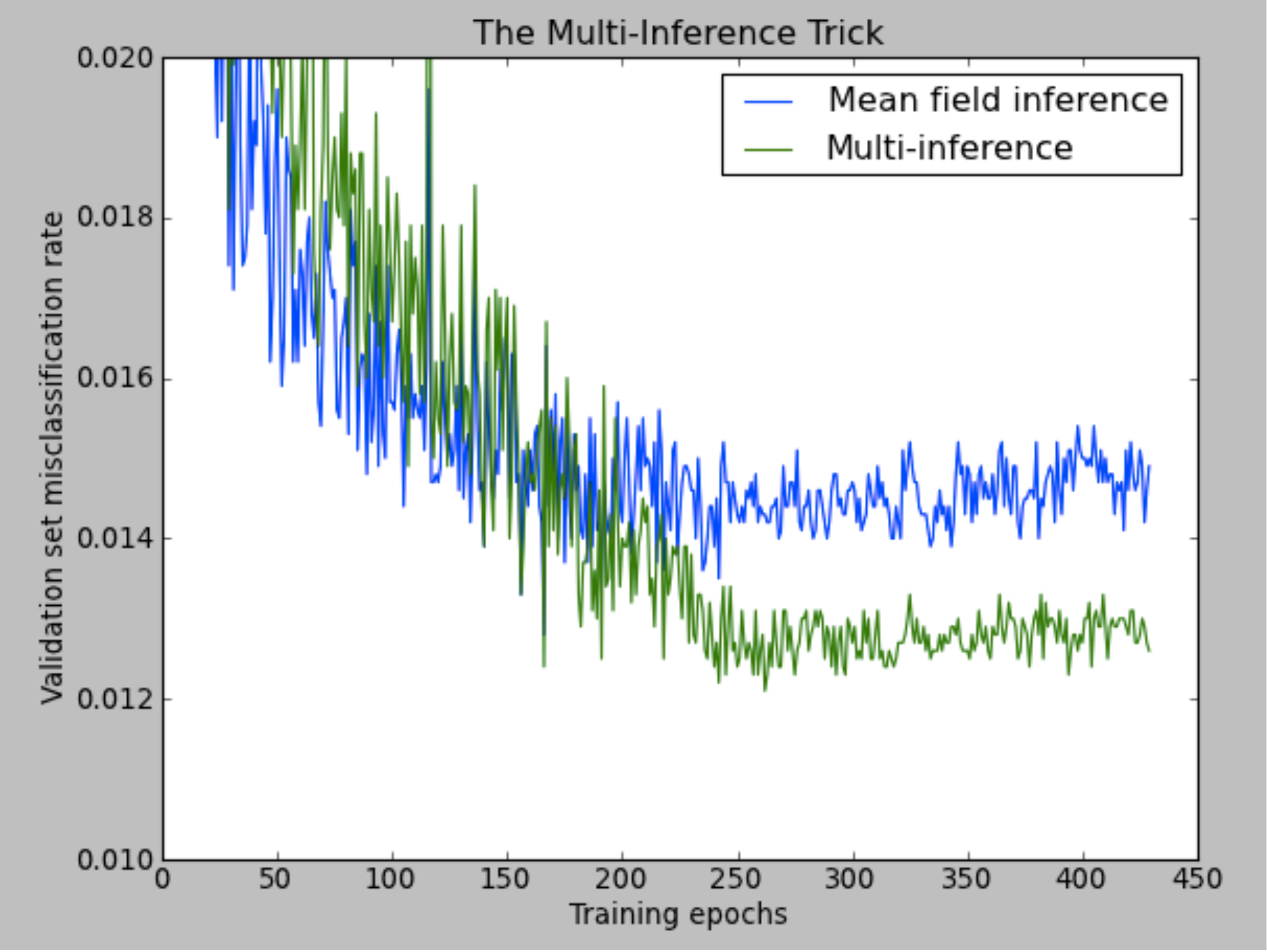}
\caption{
Here we use MP training on MNIST with only 5 mean field iterations, for a set of hyperparamters where 10 mean field
iterations are necessary to get good convergence. The classification error rate for mean field quits improving
after about 150 epochs of training, but the multi-inference trick allows us to extract more information from the
model. The model continues improving at multi-inference for 100 epochs after its performance at mean field inference
has stagnated.
}
\label{mi_curve}
\end{figure}

Mean field inference can be expensive due to needing to run the fixed point equations several times in order
to reach convergence. In order to reduce this computational expense, it is possible to train using fewer mean
field iterations than required to reach convergence. In this case, we are no longer necessarily minimizing $J$
as written, but rather doing partial training of a large number of fixed-iteration recurrent nets that solve
related problems.

We can approximately take the geometric mean over all predicted distributions $Q$ and renormalize in order to
combine the predictions of all of these recurrent nets. This way, imperfections in the training procedure are
averaged out, and we are able to solve inference tasks even if the corresponding recurrent net was never sampled
during MP training.

In order to approximate this average efficiently, we simply take the geometric mean at each step of inference,
instead of attempting to take the correct geometric mean of the entire inference process. This is the same type
of approximation used to take the average over several MLP predictions when using dropout \citep{Hinton-et-al-arxiv2012}.
Here, the averaging rule is slightly different. In dropout, the different MLPs we average over either include or
exclude diferent each variable. To take the geometric mean over a unit $h_j$ that receives input from $v_i$, we
average together the contribution $v_i W_{ij}$ from the model that contains $v_i$ and the contribution $0$ from
the model that does not. The final contribution from $v_i$ is $0.5 v_i W_{ij}$ so the dropout model averaging rule
is to run an MLP with the weights divided by 2.

For the multi-inference trick, each model we are averaging over solves a different inference problem. In half the
problems, $v_i$ is observed, and constributes $v_i W_{ij}$ to $h_{j}$'s total input. In the other half of the problems,
$v_i$ is inferred. If we represent the mean field estimate of $v_i$ with $r_i$, then in this case that unit contributes
$r_i W_{ij}$ to $h_{j}$'s total input. To run multi-inference, we thus replace references to $v$ with $0.5 (v+r)$, where
$r$ is updated at each mean field iteration.

The main reason this approach is effective is that it gives a good way to incorporate information from many recurrent nets
trained in slightly different ways. However, it can also be understand as including an input denoising step built into
the inference.

See Fig. \ref{mi} for a graphical depiction of the method, and Fig. \ref{mi_curve} for an example of it in action.

\section{Justification and advantages}

In the case where we run the recurrent net for predicting $Q$ to convergence, the multi-prediction training
algorithm follows the gradient of the objective function $J$. This can be viewed as a mean field approximation
to the generalized pseudolikelihood.

While both pseudolikelihood and likelihood are asymptotically consistent estimators, their
behavior in the limited data case is different. Maximum likelihood should be better if the overall goal is to
draw realistic 
samples from the model, but generalized pseudolikelihood can often be better for training a model to answer
queries conditioning on sets similar to the $S_i$ used during training.

Note that our variational approximation is not quite the same as the way variational approximations are usually
applied. We use variational inference to ensure that the distributions we shape using backprop are as close as
possible to the true conditionals. This is different from the usual approach to variational learning, where
$Q$ is used to define a lower bound on the log likelihood and variational inference is used to make the bound
as tight as possible.

In the case where the recurrent net is not trained to convergence, there is an alternate way to justify MP training.
Rather than doing variational learning on a single probabilistic model, the MP procedure trains a family of
recurrent nets to solve related prediction problems by running for some fixed number of iterations. Each recurrent
net is trained only a subset of the data (and most recurrent nets are never trained at all, but only work because
they share parameters with the others). In this case, the multi-inference trick allows us to justify MP training as
approximately training an ensemble of recurrent nets using bagging.

\citet{Stoyanov2011} have observed that a similar training strategy is useful because it trains the model to
work well with the inference approximations it will be evaluated with at test time. We find these properties to
be useful as well. The choice of this type of variational learning combined with the underlying generalized
pseudolikelihood objective makes an MP-DBM very well suited for solving approximate inference problems but not
very well suited for sampling.

Our primary design consideration when developing multi-prediction training was ensuring that the
learning rule was state-free. PCD training uses persistent Markov chains to estimate the gradient. These Markov
chains are used to approximately sample from the model, and only sample from approximately the right distribution
if the model parameters evolve slowly. The MP training rule does not make any reference to earlier training steps,
and can be computed with no burn in. This means that the accuracy of the MP gradient is not dependent on properties
of the training algorithm such as the learning rate--properties which can easily break PCD for many choices of the
hyperparameters.

Another benefit of MP
 is that it is easy to obtain an unbiased estimate of the MP objective from a small number of samples of $v$ and $i$.
This is in contrast to the log likelihood, which requires estimating the log partition function. The best known method
for doing so is AIS, which is relatively expensive \citep{Neal-2001}. Cheap estimates of the objective function enable
early stopping based on the MP-objective (though we generally use early stopping based on classification accuracy) and
optimization based on line searches (though we do not explore that possibility in this paper).

\section{Regularization}

In order to obtain good generalization performance, \citet{Salakhutdinov2009} regularized both the weights and the
activations of the network.

\citet{Salakhutdinov2009} regularize the weights using an L2 penalty. We find that for joint training, it is critically
important not to do this. When the second layer weights are not trained well enough for them to be useful for modeling
the data, the weight decay term will drive them to become very small, and they will never have an opportunity to recover.
It is much better to use constraints on the norms of the columns of the weight vectors, as advocated by
\citet{Hinton-et-al-arxiv2012}.

\citet{Salakhutdinov2009} regularize the activities of the hidden units with a somewhat complicated sparsity penalty.
See \texttt{http://www.mit.edu/{\textasciitilde}rsalakhu/DBM.html} for details.
We use $\text{max}(|\mathbb{E}_{h\sim Q(h)} [h] - t| - \lambda, 0)$
and backpropagate this through the entire inference graph.
$t$ and $\lambda$ are hyperparameters.

\section{Related work: centering}

\citet{Montavon2012arxiv} showed that reparameterizing the DBM to improve the condition number of the Hessian results in succesful generative training without a greedy layerwise pretraining step. However, this method has never been shown to have
good classification performance, possibly because the reparameterization makes the features never be zero from the point of view of the final classifier.

We evaluate its classification performance in more detail in this work. We consider two methods of PCD training. In one,
we use Rao-Blackwellization of the negative phase particles to reduce the variance of the negative phase. In the other
variant, we use a special negative phase that \citet{Salakhutdinov2009} found useful. This negative phase uses a small
amount of mean field, which reduces the variance further but introduces some bias, and has better symmetry with the
positive phase. See \texttt{http://www.mit.edu/{\textasciitilde}rsalakhu/DBM.html} for details.

\section{MNIST experiments}

\begin{figure}
\center
\includegraphics[width=0.5 \textwidth]{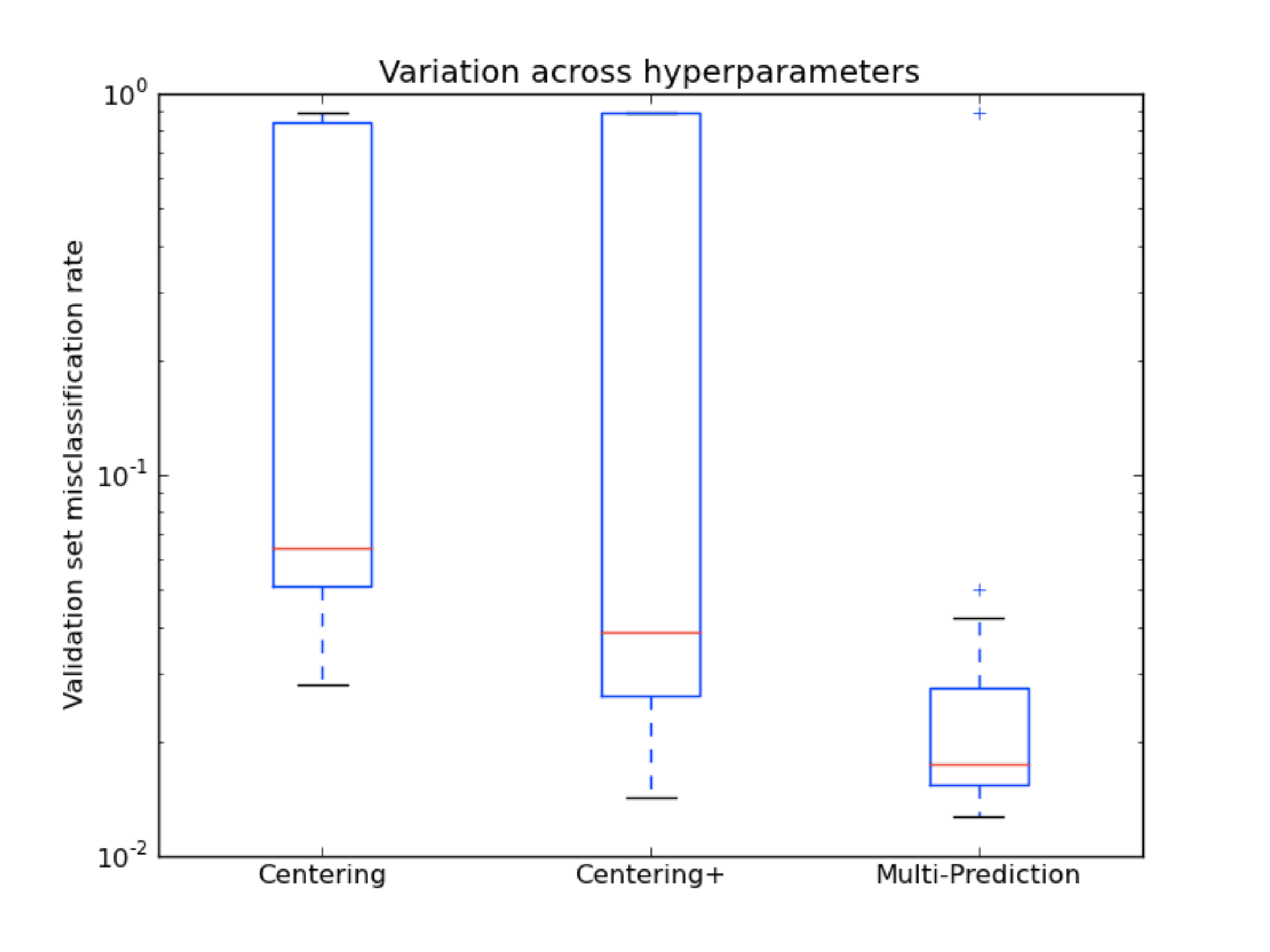}
\caption{
During cross-validation, MP training consistently performs better at classification than either centering or
centering with the special negative phase.
}
\label{crossval}
\end{figure}

\begin{figure}
\center
\includegraphics[width=0.5 \textwidth]{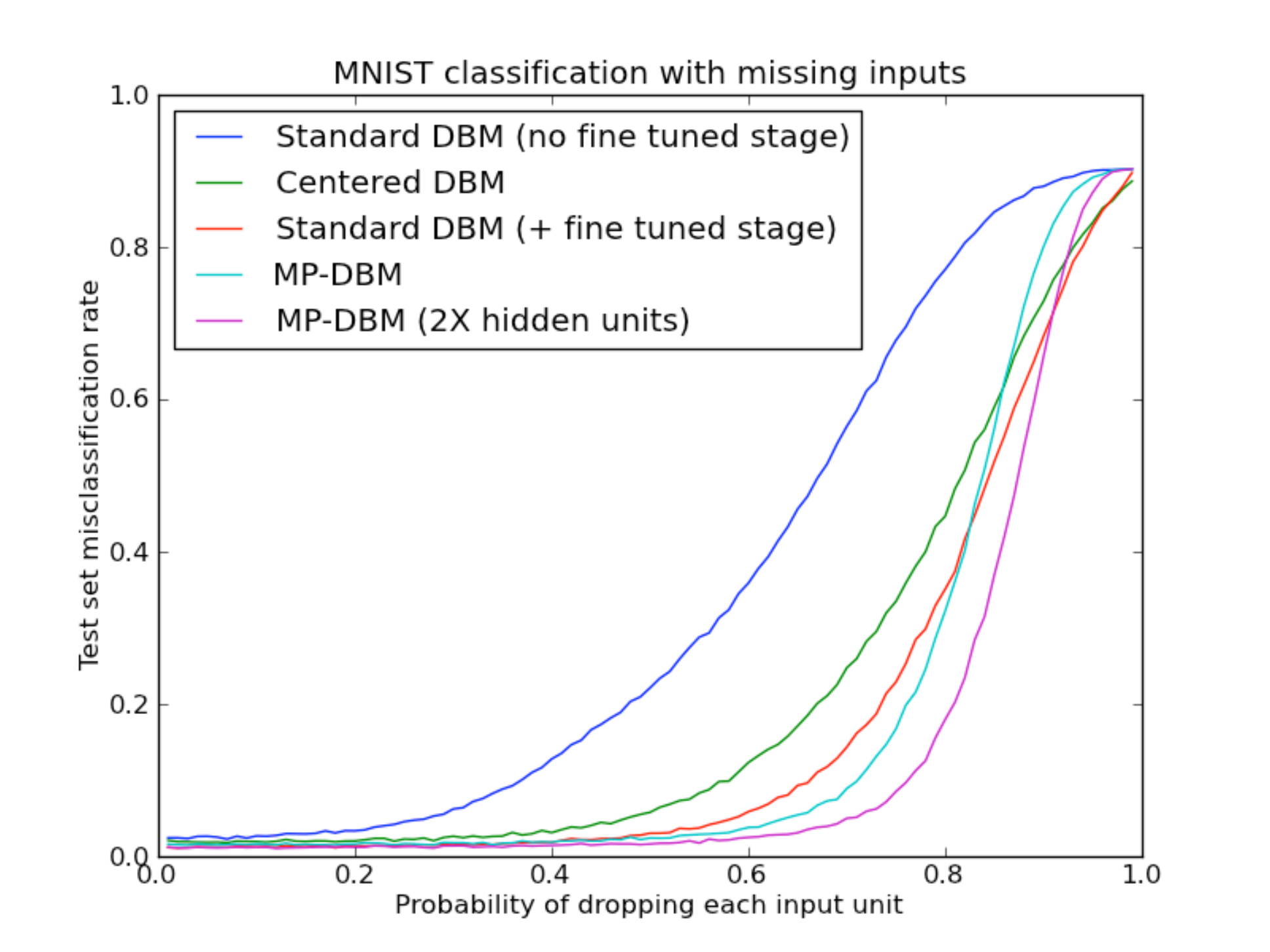}
\caption{
When classifying with missing inputs, the MP-DBM outperforms the other DBMs for most amounts of missing inputs.
}
\label{missing_inputs}
\end{figure}

\begin{figure}
\center
\includegraphics[width=0.5 \textwidth]{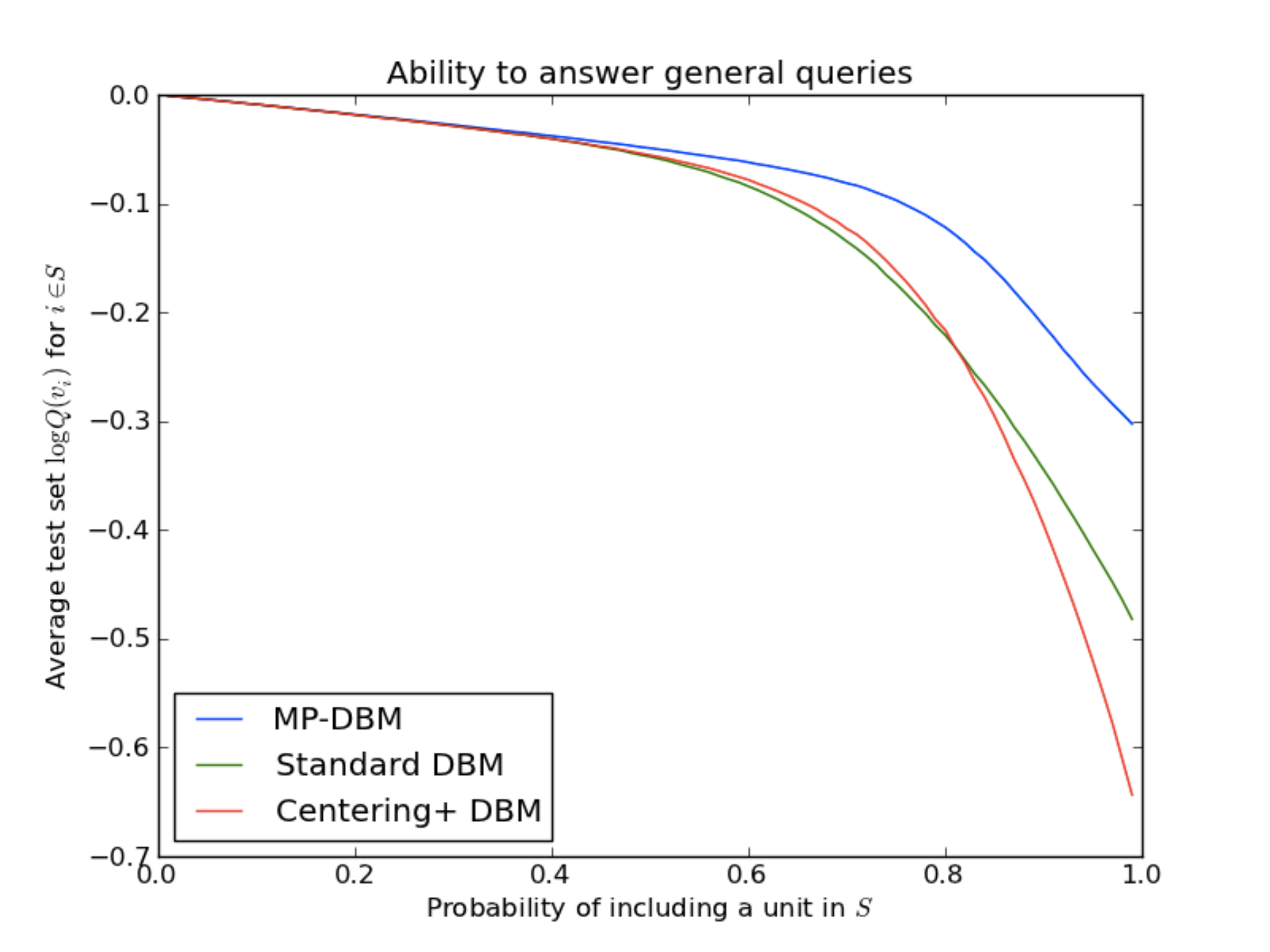}
\caption{
When using approximate inference to resolve general queries, the standard DBM, centered DBM, and MP-DBM all perform
about the same when asked to predict a small number of variables. For larger queries, the MP-DBM performs the best.
}
\label{general_queries}
\end{figure}

In order to compare MP training and centering to standard DBM performance, we cross-validated each of the new methods
by running 25 training experiments for each of three conditions:  centered DBMs, centered DBMs
with the special negative phase (``Centering+''), and MP training. For these experiments we did not use the multi-inference
trick.

All three conditions visited exactly the same set of 25 hyperparameter values for the momentum schedule, sparsity
regularization hyperparameters, weight and bias initialization hyperparameters, weight norm constraint values, and
number of mean field iterations. The centered DBMs also required one additional hyperparameter, the number of Gibbs
steps to run for PCD.

We used different values of the learning rate for the different conditions, because the different conditions require
different ranges of learning rate to perform well.

We use the same size of model, minibatch and negative chain collection as \citet{Salakhutdinov2009},
with 500 hidden units in the first layer, 1,000 hidden units in the second,
100 examples per minibatch, and 100 negative chains.

See Fig. \ref{crossval} for the results of cross-validation. On the validation set, MP training consistently performs
better and is much less sensitive to hyperparameters than the other methods. This is likely because the state-free nature
of the learning rule makes it perform better with settings of the learning rate and momentum schedule that result in the
model distribution changing too fast for a method based on Markov chains to keep up.

When we fine-tune the best model, the best ``Centering+'' DBM obtains a classification error of 1.22 \% on the test set.
The best MP-DBM obtains a classification error of 0.99 \%. This compares to 0.95 \% obtained by \citet{Salakhutdinov2009}.

If instead of adding an MLP to the model to do fine tuning, we simply train a larger MP-DBM with twice as many hidden units
in each layer, and apply the multi-inference trick, we obtain a slightly better classification error rate of 0.91 \%.
In other words, we are able to classify better using a single large DBM and a generic inference procedure, rather than using
a DBM followed by an entirely separate MLP model specalized for classification.

The original DBM was motivated primarily as a generative model with a high AIS score and as a classifier. Here we explore some
more uses of the DBM as a generative model.
First, we evaluate the use of the DBM to classify in the presence of missing inputs. See Fig. \ref{missing_inputs} for details. We find that for most
amounts of missing inputs, the MP-DBM classifies better than the standard DBM or the best centering DBM. We also explored the
ability of the DBM to resolve queries about random subsets of variables. See Fig. \ref{general_queries} for details. Again, we find
that the MP-DBM outperforms the other DBMs.

\section{Conclusion}

This paper has demonstrated that MP training and the multi-inference trick provide a means of training a single model, with
a single stage of training, that matches the performance of standard DBMs but still works as a general probabilistic model,
capable of handling missing inputs and answering general queries. In future work, we hope to obtain state of the art performance
by combining MP training with dropout, and also to apply this method to other datasets.

\bibliography{strings,strings-shorter,ml,aigaion-shorter,aigaion}
\bibliographystyle{natbib}

\end{document}